Please cite this paper as:

S. Garg, N. Suenderhauf, and M. Milford, "Don't Look Back: Robustifying Place Categorization for Viewpoint- and Condition-Invariant Place Recognition," in IEEE International Conference on Robotics and Automation (ICRA), 2018.

bibtex:

```
@inproceedings{garg2018don't,
  title={Don't Look Back: Robustifying Place Categorization for Viewpoint- and Condition-Invariant Place Recognition},
  author={Garg, Sourav and Suenderhauf, Niko and Milford, Michael},
  booktitle={IEEE International Conference on Robotics and Automation (ICRA)},
  year={2018}
}
```

# Don't Look Back: Robustifying Place Categorization for Viewpoint- and Condition-Invariant Place Recognition

Sourav Garg, Niko Suenderhauf and Michael Milford

*Abstract*— When a human drives a car along a road for the first time, they later recognize where they are on the *return* journey typically without needing to look in their rear view mirror or turn around to look back, despite significant viewpoint and appearance change. Such navigation capabilities are typically attributed to our *semantic* visual understanding of the environment [1] beyond geometry to recognizing the types of places we are passing through such as "passing a shop on the left" or "moving through a forested area". Humans are in effect using place *categorization* [2] to perform *specific* place recognition even when the viewpoint is 180 degrees reversed. Recent advances in deep neural networks have enabled high performance semantic understanding of visual places and scenes, opening up the possibility of emulating what humans do. In this work, we develop a novel methodology for using the semantics-aware higher-order layers of deep neural networks for recognizing *specific* places from within a reference database. To further improve the robustness to appearance change, we develop a descriptor normalization scheme that builds on the success of normalization schemes for pure appearance-based techniques such as SeqSLAM [3]. Using two different datasets — one road-based, one pedestrian-based, we evaluate the performance of the system in performing place recognition on reverse traversals of a route with a limited field of view camera and no turn-back-and-look behaviours, and compare to existing state-of-the-art techniques and vanilla off-the-shelf features. The results demonstrate significant improvements over the existing state of the art, especially for extreme perceptual challenges that involve both great viewpoint change *and* environmental appearance change. We also provide experimental analyses of the contributions of the various system components: the use of spatio-temporal sequences, place categorization and place-centric characteristics as opposed to object-centric semantics.

## I. INTRODUCTION

Humans interpret scenes through the visual semantics or *gist* [4] of the visual information. The theory of processing increasingly complex visual components (edges, shapes, objects, and scene) in a hierarchical manner [5], now practically possible using deep neural networks [6], shows that the later components in hierarchy provide access to the meaning of the scene. The *geon* theory [5] also establishes the viewpoint-invariant understanding of visual information, as also demonstrated by the higher-order layers of convolutional neural networks for semantic understanding of visual places [7].

The task of visual place categorization or *generic* place recognition can be performed by directly using the pre-trained deep neural networks [2]. But, recognizing *specific*

The authors are with the Australian Centre for Robotic Vision at the Queensland University of Technology. This work was supported by an ARC Centre of Excellence for Robotic Vision Grant CE140100016. SG is supported by QUT Postgraduate Research Award. MM is supported by an Australian Research Council Future Fellowship FT140101229.

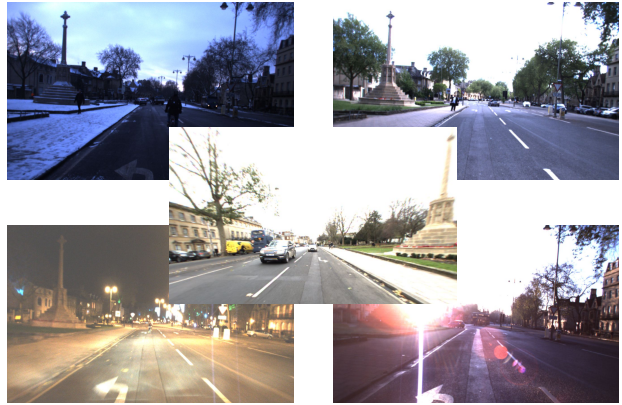

Fig. 1. Rear-View (center) and front-view images from different traverses of the Oxford Dataset [8]. Visual place recognition for viewpoint variations as extreme as front versus rear view imagery, under changing environmental conditions, requires semantic scene understanding.

places from within a reference set of places is not straightforward. This is ascribed to the specificity and invariance required in the representation of these places in order to determine a true match. For example, a *generic* place category like 'parking garage' does not provide sufficient information about which *specific* location within the 'parking garage' the camera is looking at. Furthermore, the problem becomes more challenging due to extreme viewpoint and appearance variations that a place may undergo when it is revisited. Fig. 1 shows sample images from Oxford Robotcar dataset [8] depicting a place that exhibits viewpoint variations with viewing direction flipped from front to rear and appearance variations due to changing environmental conditions like time of day, weather and season.

The large-scale visual place recognition methods like FAB-MAP [9] often lack robustness to vast appearance variations. The appearance-robust methods like SeqSLAM [3] are invariant to challenging environmental conditions, but at the cost of viewpoint-dependence and velocity-sensitivity. The use of hand-crafted local features like SURF [10] or global image representations like HoG [11] for visual place recognition [12], [13] respectively, is continually being replaced by deep-learned feature representations [14], [15].

In this paper, we investigate the suitability of semantics-aware higher-order fully-connected layers of deep neural networks, as opposed to viewpoint-dependent middle-order convolutional layers [7], for viewpoint- and condition-invariant place recognition. We particularly investigate from the perspective of robustifying semantic place categorization networks for place recognition, in order to retain both the capabilities. We demonstrate the effectiveness of semantics-aware representations for handling viewpoint variations as extreme as front versus rear view. Further, we propose to use descriptor normalization to enable appearance-robustness against changing environmental conditions. We also show that the contextual information from the scene, for example, the left and right image regions, can be used to create an extended image descriptor for further improving recognition performance. This is specifically beneficial for route traversals using forward-facing cameras, for example, in autonomous vehicles. Finally, we present valuable insights from the PCA analysis of the place descriptors that highlight the importance of spatio-temporal nature of the information inherent within the place recognition problem. The comparative study of neural networks trained on different types of data shows that place-centric semantics aid in recognizing places by inherently ignoring the dynamic object-level information, for example, pedestrians and vehicles on road.

The paper proceeds as follows: Section II reviews the research work related to appearance-robust visual place recognition, use of semantics, and front-rear image matching; Section III highlights the key components of our proposed approach; Section IV explains the datasets and performance measures used for experiments; Section V shows the quantitative and qualitative results obtained using the proposed methodology; Section VI discusses the sensitivity of the method; finally, Section VII concludes the paper with scope of future work. The qualitative results are also available on website [1].

## II. LITERATURE REVIEW

### A. Visual Place Recognition

Visual Place Recognition has received significant attention in robotics with development of methods based on appearance as well as geometry, and now gradually moving towards using visual semantics. The appearance-only based methods like FAB-MAP [9] use visual Bag of Words (BoW) approach to construct a visual vocabulary using robust features such as SURF [10]. The appearance-based methods are often supplemented with geometric information [12] to further improve the robustness of the system.

However, recognizing places in challenging environmental conditions such as varying season, time of day and weather conditions, is a challenging task. The methods based on sequence-searching like SeqSLAM [3], SMART [16], appearance prediction [17], shadow-removal [18], illumination-invariance [19], have been shown to work well under extreme appearance variations, but most of these methods lack robustness to viewpoint variations.

Recently, more robust place representations have been proposed using deep-learned features like ConvNet Landmarks [20], direct use of different layers of CNNs [21], [6] as feature, for example, AMOSNet [22], [7], or weakly supervised NetVLAD architecture [14]. These appearance-robust methods have been shown to work only for moderate viewpoint variations and do not use semantic information in any form.

[1]https://sites.google.com/view/robust-place-recognition

### B. Role of Semantics

The use of visual semantics is more pronounced in simultaneous localization and mapping (SLAM) than place recognition. Object-level semantic information is often utilized either in form of pre-trained 3D models [23] or as a part of pose optimization equation [24], [25]. The authors in [26] combine object recognition and semantic image segmentation for dense semantic SLAM.

However, *object-centric* approaches are often not transferable to *place-centric* environments where the visual content of the scene is not necessarily focused on objects. The deep-learned place categorization [21] network trained on *place-centric* data captures the semantic information that can be used to recognize places as a *general* category. On the other hand, the 'traditional' visual place recognition problem requires recognizing *specific* places within a reference database. The use of semantics for (specific) visual place recognition has received limited attention. The use of Semantic Bag of Words (SBoW) [27], Semantic Landmarks [28] and semantic segmentation of persistent regions [15], has been shown to improve visual place recognition by explicitly using visual semantic information. However, [27] has not been shown to be robust to appearance variations and [28], [15] use semantic information specific to road-based environments. [29] uses place categorization information only for semantic segmentation of the reference database in order to improve condition-invariant place recognition performance.

The deep-learned representations of images encapsulate task-specific visual semantics depending on the training data and task at hand [2]. The use of place-centric semantic information for visual place recognition has been emphasized in [15], [7]. [15] explicitly employs semantic masks for aggregating image descriptors to improve place recognition, but considers only road-based semantics. [7] demonstrated that higher-order fully connected layers of the convolutional networks, though viewpoint-invariant, lack appearance-robustness. Our proposed approach exploits the semantic information contained within the deep convolutional networks for enabling *both* viewpoint- and condition-invariant place recognition in different environmental settings, while maintaining the capability of the deep network to semantically categorize and recognize places simultaneously.

### C. Front vs Rear View Matching

Variations in viewpoint for previous visual place recognition research are mostly limited in extent; for example, variation in lateral displacement, orientation, and scale relative to the reference 6-DoF camera pose in the real world. Changes in viewpoint as extreme as front- vs rear-view in place recognition research have not been explicitly addressed to the best of our knowledge, though the use of disjoint field-of-view cameras has been explored in camera calibration [30], motion estimation [31], and mapping [32]. The motion estimation method in [31] uses the warped rear-view patches from buildings and roads to match with the front-view patches using a Manhattan World assumption. However, place recognition using a similar approach will require appearance-robust patch description. The authors in [33] adapt localization sensory window for an out-of-order place recognition but the viewpoint remains the same across different segments of the route traversal.

The geometric constraints within an image captured from the front-view do not match with those captured from the rear-view, as also established in [31], where the authors show image matching failure using SURF-based homography for opposite viewpoints of the scene. Therefore, this problem requires human-like semantic understanding of the scene to be able to match places, even under significant appearance variations in the environment, which sets the premise of our work.

## III. PROPOSED APPROACH

A robust place representation is a vital part of a visual place recognition pipeline. Our proposed approach builds on the success of deep-learned place representations, leveraging the semantics-aware higher-order layer features. A key component of our approach is to use descriptor normalization that immensely improves the robustness of these viewpoint-invariant features to changing environmental conditions, like day versus night. We also propose an extension of these feature representations created by concatenating the descriptors obtained from left and right portions of the images as shown in Fig. 2. Place recognition is performed by computing cosine distance between query images and the reference database to form a cost matrix, which is then searched for matching sequence of images.

### A. Place Representation

We use the state-of-the-art place categorization CNN Places365 [2], trained on place-centric data as opposed to object-centric object recognition networks [6], for representing places. The higher-order fully connected layers of the network encode a semantic description of the place [7]. We use the 'fc6' layer for this purpose which is of 4096 dimensions. Although it has been demonstrated that higher-order layers are invariant to viewpoint variations, they lack appearance-robustness [7]. In order to increase the robustness to such variations, we propose a feature normalization method as described in the subsequent section.

### B. Feature Normalization

The use of cosine or Euclidean distance for feature matching assigns different weights to feature dimensions depending on their scale range [34]. The use of feature normalization techniques for speech recognition [35] and image retrieval [34] has been shown to be useful as it increases the discrimination capabilities of the distance metric. Along the same lines, we apply the following normalization operation

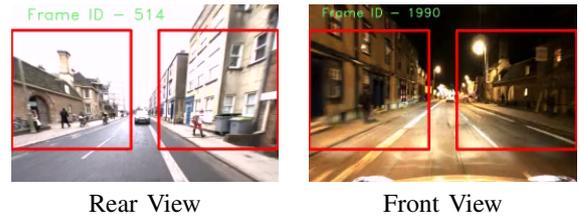

Fig. 2. Features extracted from cropped parts (marked with red boxes) of the images are concatenated together to further improve place recognition performance.

on the 'fc6' layer feature descriptors, referred to as $f_i$, for a given image $i$ in the database:

$$f'_i = \frac{(f_i - \mu_s)}{\sigma_s} \quad \forall i \qquad (1)$$

where, $\mu_s$ and $\sigma_s$ represent the mean and standard deviation of the feature descriptors computed over the entire set of images within the database. The resulting set of descriptors $f'_i$ is hereby referred to as the *N*ormalized *S*et of *D*escriptors (NSD). The dimension of each of the $f_i$, $f'_i$, $\mu_s$ and $\sigma_s$ is the same as the number of units in the layer, that is, 4096 for 'fc6'. The reference database, available beforehand, is normalized using all the images within the database. The query database is processed in an online manner as the images become available during the traverse, which means $\mu_s$ and $\sigma_s$ for the query database are updated with every new query image.

### C. Sequence Search in Cost Matrix

The query descriptor is matched with each of the reference descriptors using cosine distance to form a cost matrix. The place recognition matches are then searched in the cost matrix using a sequence matching method as described in SeqSLAM [3] and briefly as following:

$$S_i = \min_k \sum_{t=T-l}^{T} (D_k^t) \qquad (2)$$

where $S_i$ is the minimum sequence cost with respect to the reference image $i$. $k$ is the slope at which a sequence is searched and is varied within $\pm 0.2$ rad of the cost matrix diagonal. $D^T$ is the cosine distance at current time $T$, which is accumulated over a sequence of length $l$.

$$I_{min} = \arg\min_i S_i \quad \forall i \in N \qquad (3)$$

where $I_{min}$ is the matching reference image obtained by finding the lowest cost sequence among the set of $N$ images in the reference database.

### D. Cropped Regions

The route-based traversals of the environment using forward-facing cameras, for example, in autonomous vehicles, possess useful information mostly within the left and right regions of the image, as also demonstrated qualitatively in [15]. In order to further improve the performance of the proposed approach for matching, we use an extended descriptor by cropping the left and right portions of the images as shown in Fig. 2. The feature extraction and normalization is done separately for both the left and right portions and the obtained feature descriptors are concatenated together to form a $8192$ dimensional descriptor. Instead of assuming the order for concatenating the descriptors to be opposite for reference (e.g. rear-view) and query databases (e.g. front-view), we fix the order of concatenation for reference to be *left-right*. For query images, we use both the *left-right* and *right-left* order and then select the minimum of the two cosine distances for the cost matrix. We refer to this modified approach as NSD-CR in subsequent sections.

## IV. EXPERIMENTAL SETUP

### A. Datasets

We used two datasets for the experiments as described in the following subsections. While one of the datasets is road-based and exhibits only few place semantic categories, the second dataset is a pedestrian-based campus environment exhibiting diverse semantic place categories. The aerial view of trajectories is as shown in Fig. 3.

*1) Oxford Robotcar:* The Oxford Robotcar Dataset [8] comprises traverses of Oxford city during different seasons, time of day and weather conditions, capturing images using cameras pointing in all four directions. We used an initial 2.5 km traverse from 'stereo/left' and 'mono/rear' camera views for different environmental conditions, that is, Overcast Autumn (2014-12-09-13-21-02), Night Autumn (2014-12-10-18-10-50), Sunny Autumn (2014-12-16-09-14-09), Overcast Winter (2015-02-03-08-45-10) and Overcast Summer (2015-05-19-14-06-38). The sample images from these datasets are shown in Fig. 1. The ground truth matches were generated using the GPS data. We further used the GPS data to sample image frames at a constant distance of approximately 2 meters.

*2) University Campus:* The University Campus dataset was collected using a hand-held mobile phone camera by walking through the QUT Campus, traversing diverse place categories, for example, parking area, campus, corridor, botanical garden, food court, alley, crosswalk etc. as shown in Fig. 4. Therefore it covers different types of environments like indoor vs outdoor and man-made vs natural. The dataset comprises 3 traverses of 1 km each, namely: Forward-Day, Forward-Night, Reverse-Day. The backward and forward traverses refer to the same route traversed in opposite directions.

### B. Ground Truth & Peformance Measure

The ground truth matches for the Campus dataset were generated manually for 20 intermittent locations within the traverse and then interpolated for the entire traverse. For the Oxford dataset, GPS information was used to associate places from the same physical location. A match is considered a true positive if it lies within a range of its ground truth: 20m for University Campus and 40m for Oxford Robotcar dataset, a thresholding similar to [9], [3], [16]. We use the trajectory uniqueness threshold as described in [3] to

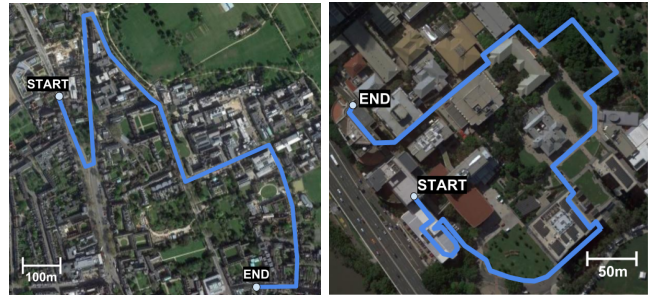

Fig. 3. Aerial view of ground truth trajectories for Oxford Robotcar (left) and University Campus (right) dataset. Source: Google Map

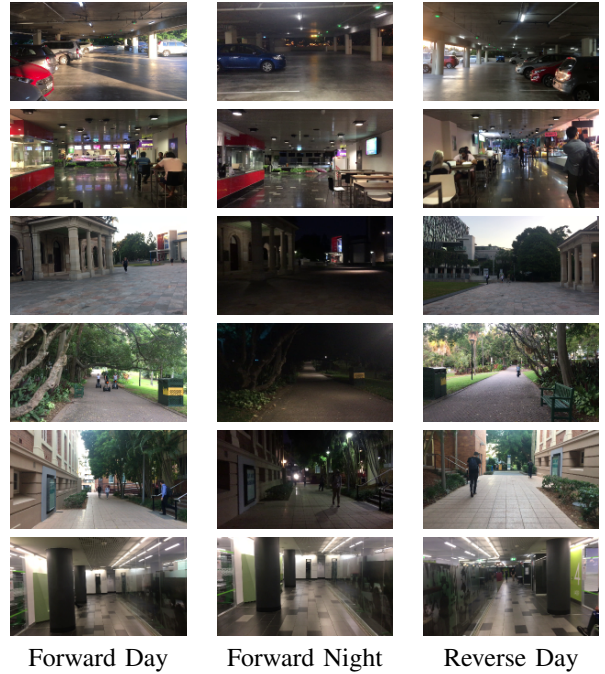

Forward Day     Forward Night     Reverse Day

Fig. 4. Sample images from the University Campus dataset showing diverse categories of places within the traverse form top to bottom, namely, parking garage, food court, campus, botanical garden, alley, and elevator lobby, as categorized by Places365 CNN.

generate the precision-recall curves and the max-F1 score which is used as a performance measure for comparative analysis.

### C. Performance Comparison

The performance is compared for the following scenarios:

- Raw-conv3: Off-the-shelf whole-image features from Place365 AlexNet CNN [21] using 'conv3' layer.
- Raw-fc6: Off-the-shelf whole-image features from Places365 AlexNet CNN using 'fc6' layer.
- NSD-fc6: Our proposed normalization operation on Raw-fc6 features.
- NSD-CR: Our proposed extended description based on cropped regions using NSD-fc6 features.
- SeqSLAM [3]: State-of-the-art condition-invariant place recognition method.

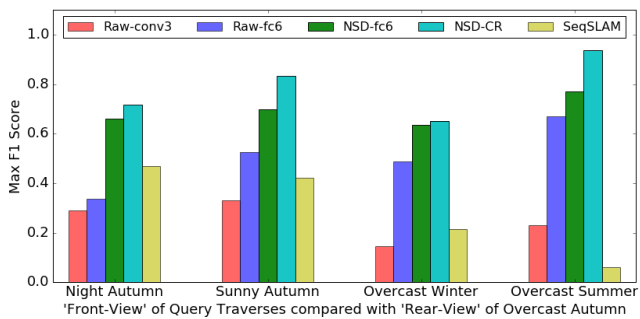

Fig. 5. *Front- vs Rear-View* results for varying environmental conditions in Oxford Robotcar dataset. Our proposed approach NSD-fc6 performs better than other methods; NSD-CR further improves the performance.

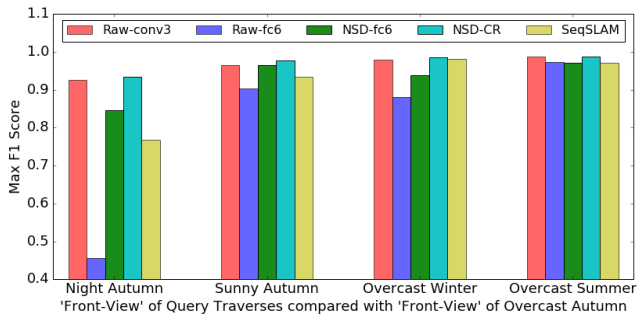

Fig. 6. *Front- vs Front-View* results for varying environmental conditions in Oxford Robotcar dataset. Raw-conv3 works better than Raw-fc6 for this case due to its high condition-invariance under limited viewpoint variations. Further, NSD-fc6 helps improve performance for Raw-fc6; and NSD-CR performs the best amongst all.

The off-the-shelf deep-learned feature representations, especially 'conv-3', have been used for place recognition by authors in [36], [37], [15], [22], [7].

## V. RESULTS

### A. Performance Across Datasets

*1) Oxford Robtocar:* We use the rear-view imagery from the Overcast Autumn traverse as the reference database and match it with the front-view imagery of all the five traverses as query databases, exhibiting different environmental conditions as shown in Fig. 1. The performance study shown in Fig. 5 shows significant improvement attained using the proposed approach as compared to the raw off-the-shelf descriptor matching. Also, using the cropped regions (NSD-CR) further boosts the performance. The Overcast Autumn traverse has the highest performance because the query and reference imagery belong to the same traverse and exhibit no change in appearance of environment, though the viewpoint is opposite for both. Moreover, the local traffic remains the same for this particular scenario, unlike other traverses where pedestrians and vehicles do not provide any useful information for recognizing places. The sequence length used for these experiments was approximately 80 meters.

*2) University Campus:* We used the Forward Night (FN) traverse as the reference database and other two: Forward

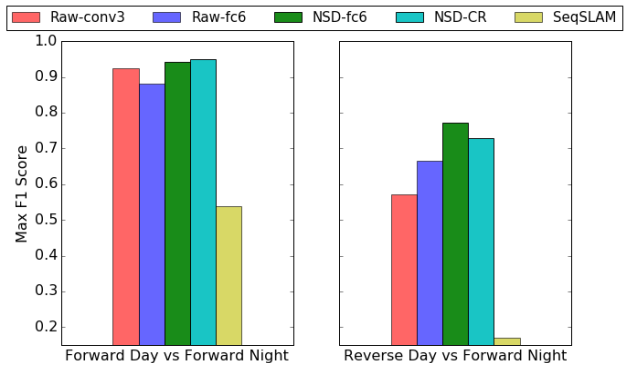

Fig. 7. University Campus: The *Forward-Forward* comparison (left) shows that performance using NSD-fc6 is slightly better than Raw-conv3. The *Reverse-Forward* comparison (right) shows 35% improvement in max F-score using NSD-fc6 (0.77) as compared to Raw-conv3 (0.57).

Day (FD) and Reverse Day (RD) as query databases. The first comparison tests the condition-invariance only and the second comparison tests both condition- and viewpoint-invariance as shown in Fig. 7. Though there is consistent performance improvement using the proposed approach, the overall absolute performance for FN-RD is low as compared to FN-FD because of extreme variations in *both* viewpoint as well as environmental conditions. The sequence length of 20 meters was used for all the five comparisons using forward-forward (FN-FD) traverses and 60 meters for reverse-forward (FN-RD) traverses.

### B. Performance Across Layers

The middle-order convolutional layers of the CNN maintain the spatial structure of the input image and have been proven to be more useful for visual place recognition under extreme appearance variations as compared to the other layers of the network, as established in [7]. On the contrary, the higher-order fully-connected layers capture the visual semantics and are more robust to viewpoint variations. In order to match places with an opposite viewpoint, the semantics-aware higher-order layers, like 'fc6', are therefore more useful than the middle-order layers, like 'conv3'. We limited our choice of layers for this comparison as a detailed analysis is available in [7]; we show here the performance differences due to layers that are different because they conceptually encode very different visual information. Fig. 8 shows that the proposed approach (NSD) applied to features from different layers of the network significantly improves the performance. Furthermore, it shows that features based on viewpoint-dependent 'conv3'-like layers, as used in [15], [36], [37], [22] cannot be used for scenarios where appearance and viewpoint *both* have significantly changed.

### C. Performance Across Networks

We investigated the performance effects of using the 'fc6' layer features for three differently trained AlexNet networks: 1) p365, trained only on Places365 data [2], 2) objects, trained on ImageNet [6] and 3) hybrid, trained on combined

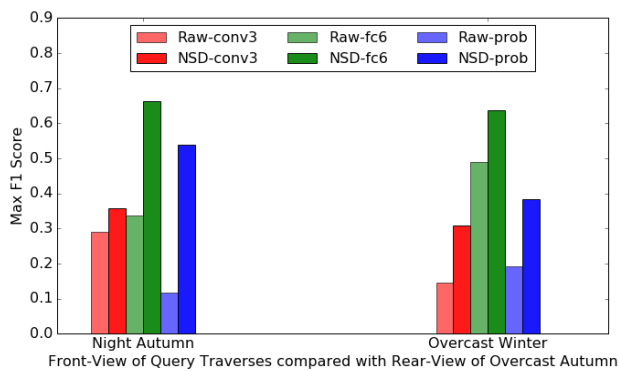

Fig. 8. Layer Comparison: The proposed approach (NSD) applied to features from conceptually different layers of the network significantly improves the performance. Furthermore, the features based on viewpoint-dependent 'conv3'-like layers are not useful for scenarios where appearance and viewpoint *both* have significantly changed.

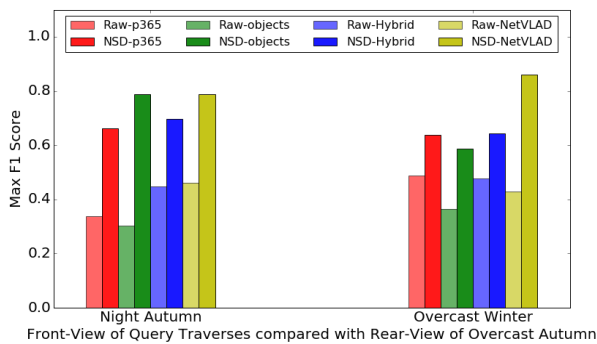

Fig. 9. Network Comparison: The proposed approach (NSD) applied to off-the-shelf features from different networks significantly improves the performance.

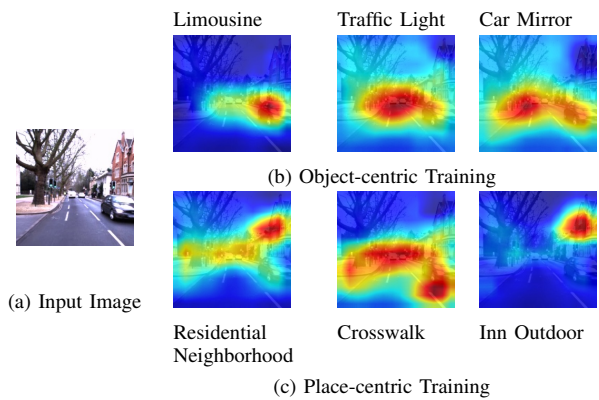

Fig. 10. Top-3 labels for an (a) input image with their class activation maps [38] for different CNNs: (b) Object-centric Training and (c) Place-centric Training. It shows that different semantic units get activated within different regions of the image depending on the type of training data (place-centric vs object-centric) used. (Image from Oxford Overcast Autumn Traverse).

Places and ImageNet data [2]. We also compared performance with NetVLAD [14] (their best performing VGG-16 + NetVLAD + whitening, Pittsburgh), which is trained specifically for the place recognition task. Fig. 9 shows that off-the-shelf descriptors from all the networks fall short on performance due to extreme variation in both viewpoint and environmental conditions. While there is a significant improvement in all the networks using our proposed approach (NSD-*), NetVLAD is benefited the most. This can be attributed either to its local feature aggregation or place-centric training, however it lacks semantic place categorization in its present form. Our choice of p365 is mainly driven by the primary goal of enabling both categorical and specific place recognition and Fig. 9 shows a consistency in performance improvement irrespective of the choice of network using the proposed approach.

The CNN trained on place-centric data like p365 and NetVLAD capture the place-centric semantic information unlike the object-centric ImageNet-based networks, as demonstrated in [21]. Therefore, place-centric training inherently selects only those parts of the image that are indicative of the category of that place. Such a selection is pertinent to visual place recognition task as demonstrated in Fig. 10, showing visualization of some of the class activation maps (CAM) of an image, generated using slightly different architectures of CNNs as proposed in [38] for both place- and object-centric data. As shown in the figure, the informative part of the image for top-3 place categories comprises patches from the buildings and the road while ignoring the vehicle on the right. On the other hand, the vehicle gets detected with highest confidence using object-centric training.

## VI. Discussion

### A. PCA Visualization

The underlying objective for performing normalization as described in Eq. 1 is to adjust the distribution of the descriptor such that it becomes more discriminative [34]. The CNNs used in this work as well as by other authors are trained for specific tasks like object recognition and place categorization. Therefore, the activations of higher-order layers, capturing the semantic information, are biased towards the semantic category to which the input image is most likely to belong. The normalization using mean and variance in Eq. 1 uniformly biases the descriptors with respect to each of its dimension such that each *specific* place within a place category can be individually identified.

The PCA visualization, similar to [39], in Fig. 11 shows a comparison between 2-D projections of raw and normalized descriptors for the Overcast Autumn traverse of Oxford Dataset. We used the final ('prob') layer for this purpose instead of 'fc6' as it is more intuitive to understand in terms of different semantic categories than in terms of different activation units of 'fc6' layer. The visualization shows that raw descriptors tend to cluster according to their semantic labels, irrespective of the image index, whereas the normalized descriptors tend to form spatio-temporal clusters, despite the absence of any explicit temporal signal during PCA training. These 2-D projections are affected only slightly for the initial few images when descriptors are normalized in an online manner.

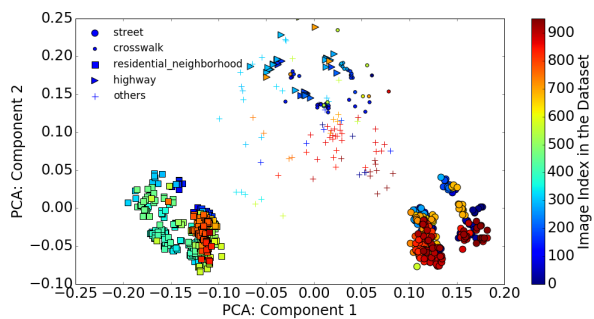

(a) Raw

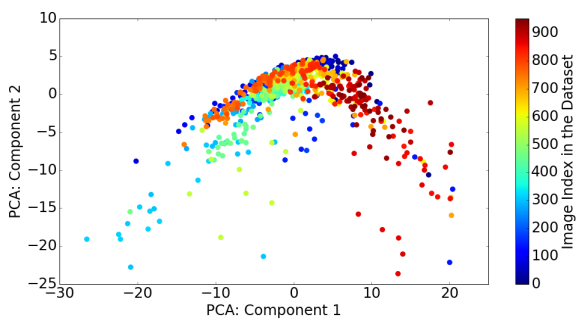

(b) Normalized

Fig. 11. PCA Visualization: The 2-D PCA projection [39] of final layer descriptors from Overcast Autumn Traverse of Oxford Dataset in Raw (top) and Normalized (bottom) form. The raw descriptors tend to cluster according to their semantic labels, irrespective of the image index, whereas, the normalized descriptors tend to form spatio-temporal clusters, despite the absence of any explicit temporal signal during PCA training.

### B. Normalization within Semantic Segments

In order to verify the efficacy of the normalization step (Eq. 1) for varying diversity within a traverse, we repeated the experiments on the University Campus dataset by performing normalization within semantic segments instead of the entire database. The semantic segments were found using the HMM-based approach described in [29] for both the reference and query databases. The segments, thus obtained, divide the database into different regions and feature normalization is performed within respective semantic segments for both the reference and query databases. Although, it might seem more convincing to use category-based segments to normalize the feature descriptors, the overall performance was similar to our proposed approach as shown in Table I.

TABLE I
PERFORMANCE VARIATION BY NORMALIZING DESCRIPTORS WITHIN SEMANTIC SEGMENTS.

| Dataset / Method | Raw | NSD | Semantic Segments |
|---|---|---|---|
| FN-FD | 0.88 | 0.94 | 0.93 |
| FN-RD | 0.66 | 0.77 | 0.76 |

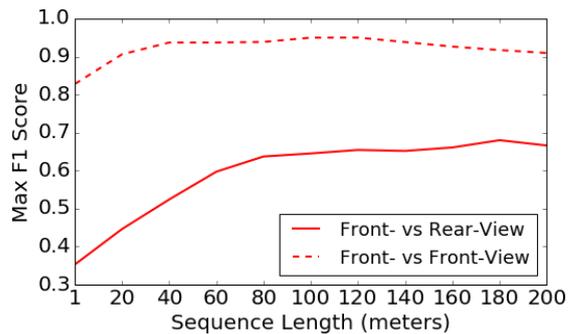

Fig. 12. The place recognition performance for extreme viewpoint variations (front- vs rear-view) benefits more from increasing sequence length as compared to the traditional front- vs front-view place matching. (Comparing Oxford Overcast Autumn and Winter Traverse).

### C. Sequence Length

In our work, we used a sequence-search approach based on SeqSLAM [3] for matching places within a cost matrix. Fig. 12 shows the effect of sequence length parameter on performance using Autumn and Winter traverse from the Oxford Robotcar dataset with for front-front and front-rear view matching. The place recognition performance for extreme viewpoint variations (front- vs rear-view) benefits more from increasing sequence length as compared to the traditional front- vs front-view place matching. Although the use of sequences for visual place recognition is appealing, attributed to the temporal structure inherent within the problem, increasing sequence length directly increases latency in the system.

## VII. CONCLUSIONS AND FUTURE WORK

We investigated the suitability of semantics-aware deep-learned feature representations from place *categorization* network for viewpoint- and condition-invariant place *recognition*. We found that the higher-order fully connected layers of the CNN, like 'fc6', pre-trained on place-centric data, exhibit invariance to viewpoint as extreme as front versus rear view of places. Further, the descriptor normalization (NSD-fc6) significantly improves the appearance-robustness of these features, therefore enabling *both* viewpoint- and condition-invariant place recognition. The extension of these feature representations (NSD-CR), obtained by concatenating the left and right portions of the image, performed the best and are especially useful for route-based place recognition. Also, the use of higher-order layer features has advantages of reduced memory footprint and computation time, attributed to the reduced dimensions of the feature vector (4k for 'fc6' as compared to 65k for 'conv3'). For the scenarios where viewpoint variations are only moderate but appearance variations are extreme, the 'conv3' features perform better as also established in [7], with further improvements attained using the proposed descriptor normalization (NSD-conv3).

The current work can be extended to a viewpoint-aware method by employing within-image semantics and performing one-to-one matching of corresponding semantic patterns.

Another possibility is to use semantic bag-of-visual-words representation of places and perform place recognition in an incremental manner with probabilistic normalization of the features with respect to a pre-learnt visual vocabulary. Finally, the appeal of the proposed system lies in the scalability and possibility of an even richer interpretation of the environment, where one could use the final layer of the deep network to categorically create a topology of the environment and then perform specific place recognition within the spatio-temporal bounds of semantic categories.

REFERENCES

[1] M. C. Potter, "Short-term conceptual memory for pictures." *Journal of experimental psychology: human learning and memory*, vol. 2, no. 5, p. 509, 1976.
[2] B. Zhou, A. Lapedriza, A. Khosla, A. Oliva, and A. Torralba, "Places: A 10 million image database for scene recognition," *IEEE Transactions on Pattern Analysis and Machine Intelligence*, 2017.
[3] M. J. Milford and G. F. Wyeth, "Seqslam: Visual route-based navigation for sunny summer days and stormy winter nights," in *Robotics and Automation (ICRA), 2012 IEEE International Conference on*. IEEE, 2012, pp. 1643–1649.
[4] A. Oliva, "Gist of the scene," *Neurobiology of attention*, vol. 696, no. 64, pp. 251–258, 2005.
[5] I. Biederman, "Recognition-by-components: a theory of human image understanding." *Psychological review*, vol. 94, no. 2, p. 115, 1987.
[6] A. Krizhevsky, I. Sutskever, and G. E. Hinton, "Imagenet classification with deep convolutional neural networks," in *Advances in neural information processing systems*, 2012, pp. 1097–1105.
[7] N. Sünderhauf, S. Shirazi, F. Dayoub, B. Upcroft, and M. Milford, "On the performance of convnet features for place recognition," in *Intelligent Robots and Systems (IROS), 2015 IEEE/RSJ International Conference on*. IEEE, 2015, pp. 4297–4304.
[8] W. Maddern, G. Pascoe, C. Linegar, and P. Newman, "1 year, 1000 km: The oxford robotcar dataset." *IJ Robotics Res.*, vol. 36, no. 1, pp. 3–15, 2017.
[9] M. Cummins and P. Newman, "Fab-map: Probabilistic localization and mapping in the space of appearance," *The International Journal of Robotics Research*, vol. 27, no. 6, pp. 647–665, 2008.
[10] H. Bay, A. Ess, T. Tuytelaars, and L. Van Gool, "Speeded-up robust features (surf)," *Computer vision and image understanding*, vol. 110, no. 3, pp. 346–359, 2008.
[11] N. Dalal and B. Triggs, "Histograms of oriented gradients for human detection," in *Computer Vision and Pattern Recognition, 2005. CVPR 2005. IEEE Computer Society Conference on*, vol. 1. IEEE, 2005, pp. 886–893.
[12] N. Kejriwal, S. Kumar, and T. Shibata, "High performance loop closure detection using bag of word pairs," *Robotics and Autonomous Systems*, vol. 77, pp. 55–65, 2016.
[13] T. Naseer, L. Spinello, W. Burgard, and C. Stachniss, "Robust visual robot localization across seasons using network flows," in *Twenty-Eighth AAAI Conference on Artificial Intelligence*, 2014.
[14] R. Arandjelovic, P. Gronat, A. Torii, T. Pajdla, and J. Sivic, "Netvlad: Cnn architecture for weakly supervised place recognition," in *Proceedings of the IEEE Conference on Computer Vision and Pattern Recognition*, 2016, pp. 5297–5307.
[15] T. Naseer, G. L. Oliveira, T. Brox, and W. Burgard, "Semantics-aware visual localization under challenging perceptual conditions," in *IEEE International Conference on Robotics and Automation (ICRA)*, 2017.
[16] E. Pepperell, P. I. Corke, and M. J. Milford, "All-environment visual place recognition with smart," in *Robotics and Automation (ICRA), 2014 IEEE International Conference on*. IEEE, 2014, pp. 1612–1618.
[17] P. Neubert, N. Sunderhauf, and P. Protzel, "Appearance change prediction for long-term navigation across seasons," in *Mobile Robots (ECMR), 2013 European Conference on*. IEEE, 2013, pp. 198–203.
[18] P. Corke, R. Paul, W. Churchill, and P. Newman, "Dealing with shadows: Capturing intrinsic scene appearance for image-based outdoor localisation," in *Intelligent Robots and Systems (IROS), 2013 IEEE/RSJ International Conference on*. IEEE, 2013, pp. 2085–2092.
[19] C. McManus, W. Churchill, W. Maddern, A. D. Stewart, and P. Newman, "Shady dealings: Robust, long-term visual localisation using illumination invariance," in *Robotics and Automation (ICRA), 2014 IEEE International Conference on*. IEEE, 2014, pp. 901–906.
[20] N. Sunderhauf, S. Shirazi, A. Jacobson, F. Dayoub, E. Pepperell, B. Upcroft, and M. Milford, "Place recognition with convnet landmarks: Viewpoint-robust, condition-robust, training-free," *Proceedings of Robotics: Science and Systems XII*, 2015.
[21] B. Zhou, A. Lapedriza, J. Xiao, A. Torralba, and A. Oliva, "Learning deep features for scene recognition using places database," in *Advances in neural information processing systems*, 2014, pp. 487–495.
[22] Z. Chen, A. Jacobson, N. Sunderhauf, B. Upcroft, L. Liu, C. Shen, I. Reid, and M. Milford, "Deep learning features at scale for visual place recognition," *arXiv preprint arXiv:1701.05105*, 2017.
[23] J. Civera, D. Gálvez-López, L. Riazuelo, J. D. Tardós, and J. Montiel, "Towards semantic slam using a monocular camera," in *Intelligent Robots and Systems (IROS), 2011 IEEE/RSJ International Conference on*. IEEE, 2011, pp. 1277–1284.
[24] R. F. Salas-Moreno, R. A. Newcombe, H. Strasdat, P. H. Kelly, and A. J. Davison, "Slam++: Simultaneous localisation and mapping at the level of objects," in *Proceedings of the IEEE conference on computer vision and pattern recognition*, 2013, pp. 1352–1359.
[25] D. Gálvez-López, M. Salas, J. D. Tardós, and J. Montiel, "Real-time monocular object slam," *Robotics and Autonomous Systems*, vol. 75, pp. 435–449, 2016.
[26] R. F. Salas-Moreno, "Dense semantic slam," Ph.D. dissertation, Imperial College London, 2014.
[27] A. Mousavian and J. Kosecka, "Semantic image based geolocation given a map," *arXiv preprint arXiv:1609.00278*, 2016.
[28] Y. Hou, H. Zhang, S. Zhou, and H. Zou, "Use of roadway scene semantic information and geometry-preserving landmark pairs to improve visual place recognition in changing environments," *IEEE Access*, 2017.
[29] S. Garg, A. Jacobson, S. Kumar, and M. Milford, "Improving condition-and environment-invariant place recognition with semantic place categorization," *arXiv preprint arXiv:1706.07144*, 2017.
[30] L. Heng, P. Furgale, and M. Pollefeys, "Leveraging image-based localization for infrastructure-based calibration of a multi-camera rig," *Journal of Field Robotics*, vol. 32, no. 5, pp. 775–802, 2015.
[31] A. Kawasaki, H. Saito, and K. Hara, "Motion estimation for non-overlapping cameras by improvement of feature points matching based on urban 3d structure," in *Image Processing (ICIP), 2015 IEEE International Conference on*. IEEE, 2015, pp. 1230–1234.
[32] M. J. Tribou, A. Harmat, D. W. Wang, I. Sharf, and S. L. Waslander, "Multi-camera parallel tracking and mapping with non-overlapping fields of view," *The International Journal of Robotics Research*, vol. 34, no. 12, pp. 1480–1500, 2015.
[33] J. Bruce, A. Jacobson, and M. Milford, "Look no further: Adapting the localization sensory window to the temporal characteristics of the environment," *IEEE Robotics and Automation Letters*, vol. 2, no. 4, pp. 2209–2216, 2017.
[34] S. Aksoy and R. M. Haralick, "Feature normalization and likelihood-based similarity measures for image retrieval," *Pattern recognition letters*, vol. 22, no. 5, pp. 563–582, 2001.
[35] A. Stolcke, S. Kajarekar, and L. Ferrer, "Nonparametric feature normalization for svm-based speaker verification," in *Acoustics, Speech and Signal Processing, 2008. ICASSP 2008. IEEE International Conference on*. IEEE, 2008, pp. 1577–1580.
[36] D. Bai, C. Wang, B. Zhang, X. Yi, and X. Yang, "Sequence searching with cnn features for robust and fast visual place recognition," *Computers & Graphics*, 2017.
[37] R. Arroyo, P. F. Alcantarilla, L. M. Bergasa, and E. Romera, "Fusion and binarization of cnn features for robust topological localization across seasons," in *Intelligent Robots and Systems (IROS), 2016 IEEE/RSJ International Conference on*. IEEE, 2016, pp. 4656–4663.
[38] B. Zhou, A. Khosla, A. Lapedriza, A. Oliva, and A. Torralba, "Learning deep features for discriminative localization," in *Proceedings of the IEEE Conference on Computer Vision and Pattern Recognition*, 2016, pp. 2921–2929.
[39] G. Patterson and J. Hays, "Sun attribute database: Discovering, annotating, and recognizing scene attributes," in *Computer Vision and Pattern Recognition (CVPR), 2012 IEEE Conference on*. IEEE, 2012, pp. 2751–2758.